\documentclass[conference]{IEEEtran}
\usepackage[T1]{fontenc}
\usepackage{cite}
\usepackage{amsmath,amssymb}
\usepackage{graphicx}
\usepackage{xcolor}
\usepackage{multirow}
\usepackage{booktabs}
\usepackage{float}
\usepackage{placeins}
\IEEEoverridecommandlockouts

\begin{document}

\title{TCINet: Tissue-Conditioned Adaptive Cross-Branch Interaction Network
for Nuclei Instance Segmentation}

\author{
\IEEEauthorblockN{
Zhenghang Qian\textsuperscript{1,\ensuremath{\dagger}},
Yang Liu\textsuperscript{1,\ensuremath{\dagger}},
Yanzhi Yu\textsuperscript{2},
Peiwu Qin\textsuperscript{3,*},
Yibing Shen\textsuperscript{1}\\
JOONGWON CHAE\textsuperscript{1},
Lihui Luo\textsuperscript{1},
Zhongde Zhang\textsuperscript{3},
Jianwen Guo\textsuperscript{3},
Lianshun Jin\textsuperscript{3}}
\IEEEauthorblockA{\textsuperscript{1}Shenzhen International Graduate School,
Tsinghua University, Shenzhen, China}
\IEEEauthorblockA{\textsuperscript{2}Macau University of Science and Technology,
Macau, China}
\IEEEauthorblockA{\textsuperscript{3}Chinese Medicine Guangdong Laboratory/Hengqin Laboratory,
Hengqin, China}
\IEEEauthorblockA{Emails:
\{qianzx21, lyang22, shenyb25, luolh23\}@mails.tsinghua.edu.cn\\
qinpeiwu@hengqinlab.cn; 306247680@qq.com; 924903534@qq.com}
\thanks{\ensuremath{\dagger}Zhenghang Qian and Yang Liu contributed equally
to this work. \textsuperscript{*}Corresponding author: Peiwu Qin
(email: qinpeiwu@hengqinlab.cn).}
}

\maketitle

\begin{abstract}
Separating touching nuclei requires local boundary detail, while
cell type classification requires broader morphological context.
These requirements also vary with the surrounding tissue environment.
We propose the Tissue-Conditioned Adaptive Cross-Branch Interaction
Network (TCINet), a single-stage model with Task-Specific Feature
Adapters (TSFA) and Tissue-Conditioned Task-Aware Cross-Branch Interaction
(TC-TACBI). TSFA produces branch-specific features at skip connections,
and TC-TACBI controls three asymmetric paths among nucleus-presence
(NP), horizontal-vertical (HV), and nucleus-type (NT) branches using
predicted tissue probabilities. On PanNuke under standard three-fold
cross-validation, TCINet achieves multiclass and binary panoptic quality
scores of 0.519 and 0.692, respectively. Zero-shot tests on MoNuSeg
yield a detection F1 score of 0.876. On PanNuke fold 0, adding TC-TACBI increases mPQ
from 0.505 to 0.525. Code is available at
https://github.com/celestial1321/TCINet.
\end{abstract}


\section{Introduction}
\label{sec:intro}

Cell morphology is shaped by the surrounding tissue environment in
hematoxylin and eosin (H\&E)-stained sections. Different tasks require
different cues: separating touching nuclei demands localized boundary
detail, while cell type classification requires broader neighborhood
context. HoVer-Net~\cite{graham2019hovernet} introduced joint distance-map
and classification prediction, CellViT~\cite{horst2024cellvit}
demonstrated the benefit of pretrained transformer encoders, and
LKCell~\cite{cui2024lkcell} widened the receptive field through large
convolutions.

Shared skip features must support all three prediction tasks. TCINet
adapts these features separately for each branch through TSFA, then
routes information along three asymmetric paths controlled by predicted
tissue probabilities through TC-TACBI. We evaluate TCINet on
PanNuke~\cite{gamper2020pannuke} and test zero-shot transfer to
MoNuSeg~\cite{kumar2020monuseg}.

\section{Related Work}
\label{sec:related}
TCINet follows CellViT's encoder and NP/HV/NT
architecture~\cite{horst2024cellvit}, with prediction heads inspired by
LKCell~\cite{cui2024lkcell}. Adapter modules~\cite{houlsby2019adapter}
and low-rank adaptation (LoRA)~\cite{hu2022lora} introduced lightweight
feature adaptation; ViT-Adapter~\cite{chen2022vitadapter} and Medical
SAM Adapter~\cite{wu2023medical} extended adaptation to vision and
medical imaging. Our TSFA operates at decoder skip connections to
produce branch-specific representations.

Cross-stitch networks and NDDR-CNN~\cite{misra2016cross,gao2019nddr}
learn cross-task feature sharing. Squeeze-and-Excitation
(SE)~\cite{hu2018squeeze} and the Convolutional Block Attention Module
(CBAM)~\cite{woo2018cbam} reweight features within a branch, while
Feature-wise Linear Modulation (FiLM)~\cite{perez2018film} applies
conditional affine transformations. TC-TACBI uses three asymmetric
cross-branch paths, with each path scaled by a tissue-type MLP.

SFE-Net~\cite{chen2024sfe} fuses spatial and frequency representations,
whereas our interaction operates on task features without frequency
transforms. PromptNucSeg-H~\cite{shui2024promptnucseg} separates prompt
generation from Segment Anything Model (SAM) segmentation.
IVAAN-H~\cite{jeong2026ivaan} uses attribute texts for vision--language
alignment, and UN-SAM~\cite{chen2025unsam} employs domain-aware
self-prompting. TCINet produces NP/HV/NT maps in one network, using
neither text encoding nor externally supplied tissue labels at inference.

\section{Method}
\label{sec:method}

\begin{figure*}[!t]
\centering
    \includegraphics[width=0.90\textwidth]{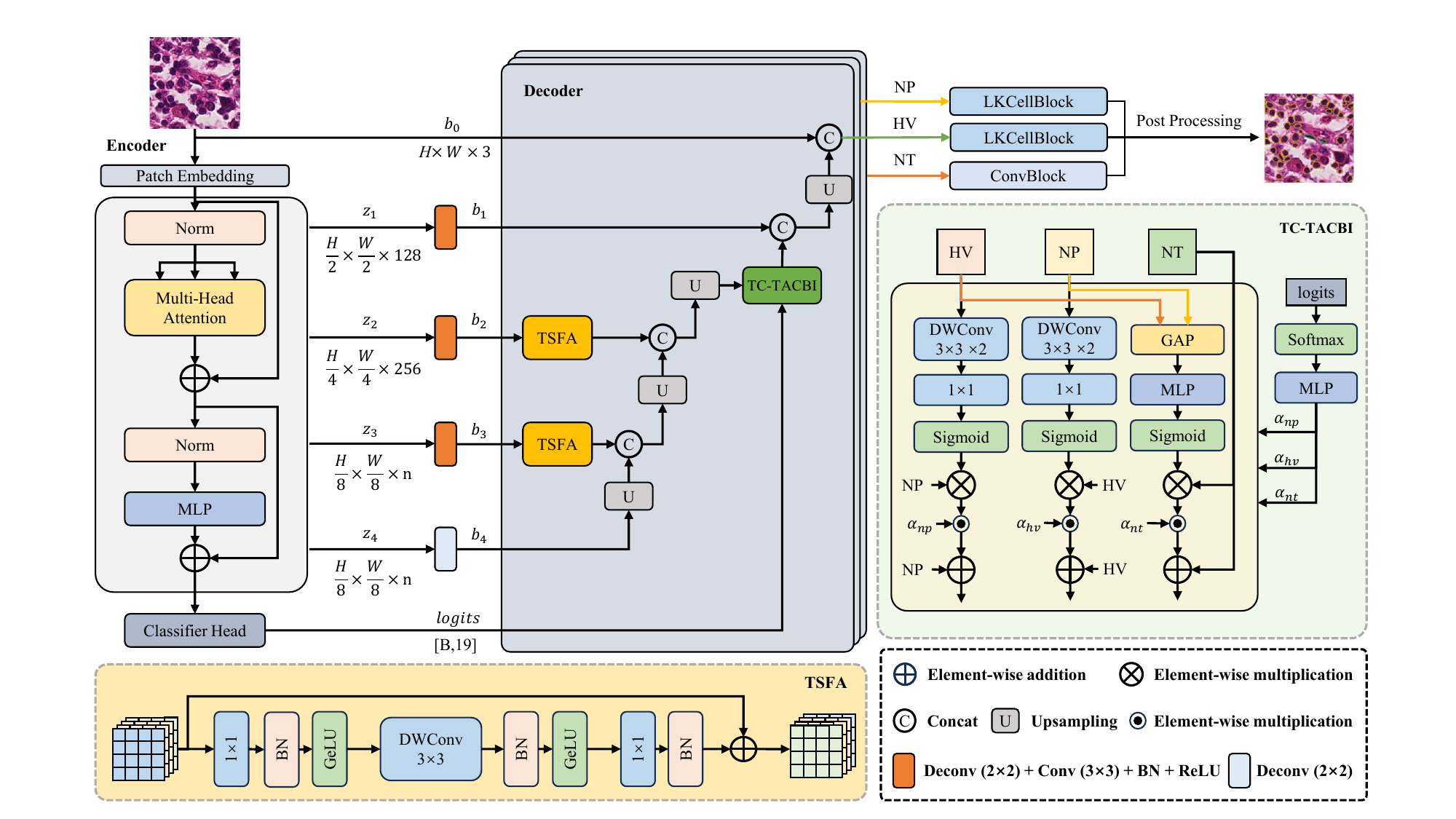}
    \caption{Architecture of TCINet. The SAM-H encoder extracts
    features and produces tissue logits. TSFA modules at the
    $b_3,b_2$ skip connections produce branch-specific features.
    TC-TACBI applies HV$\to$NP, NP$\to$HV, and (NP,HV)$\to$NT
    interactions to 128-channel branch features. Final predictions use
    LKCellBlock heads for NP and HV, and a convolutional head for NT.
    Sizes beside $z_i$ indicate decoder-rescaled features ($n=512$).}
    \label{fig:detail}
\end{figure*}

\subsection{Encoder and Task-Specific Adaptation}
TCINet uses a SAM-H vision transformer (ViT)
encoder~\cite{kirillov2023sam} (Fig.~\ref{fig:detail}). Intermediate
features $z_1$--$z_4$ are extracted at layers 8, 16, 24, and 32.
Global average pooling over the encoder-neck output feeds a linear
head that predicts 19 tissue-type logits $\mathbf{t}$. The shared
decoder produces skip features $b_3$ (512 channels) and $b_2$
(256 channels). These features pass through per-branch TSFA modules
before concatenation with upsampled branch features. TC-TACBI operates
on the resulting 128-channel branch features before final upsampling.

In branch $k$, a residual adapter reduces the channel count by half
with a $1\times1$ convolution, applies a $3\times3$ depthwise
convolution, and restores the channels with a second $1\times1$ convolution:
\begin{equation}
\begin{aligned}
H_k(F)&=\mathrm{GELU}(f_{\rm down}(F)),\\
G_k(F)&=\mathrm{GELU}(\mathrm{DW}_{3\times3}(H_k(F))),\\
\mathrm{TSFA}_k(F)&=F+f_{\rm up}(G_k(F)).
\end{aligned}
\end{equation}
Each convolution is followed by batch normalization (BN), with a
Gaussian Error Linear Unit (GELU) after the first two. The final BN
scale is zero-initialized, so TSFA starts as a pass-through and
gradually learns branch-specific adjustments. Channel reduction limits
pointwise convolution cost, while depthwise filtering processes local
morphology. The NP, HV, and NT adapters use the same shared decoder
maps but learn separate parameters.

\begin{table*}[!t]
\centering\scriptsize
\caption{Tissue-wise mPQ/bPQ averaged over three PanNuke folds.
Baseline scores and standard deviations are from the cited papers,
rounded to three decimals; missing deviations are shown as dashes.
$^{*}$Competitive recent efforts (ViT-H): IVAAN-H adds pretrained CLIP text
features and attribute-guided alignment; UN-SAM uses supplied domain
information for domain-adaptive encoding and self-prompt decoding.
TCINet performs task-specific adaptation and image-predicted tissue
conditioning in a single-stage network.}
\label{tab:pannuke_hierarchical}
\begin{tabular*}{\linewidth}{@{\extracolsep{\fill}}l cc cc cc cc cc cc@{}}
\toprule
&\multicolumn{2}{c}{CellViT-SAM-H~\cite{horst2024cellvit}}
&\multicolumn{2}{c}{LKCell-L~\cite{cui2024lkcell}}
&\multicolumn{2}{c}{PromptNucSeg-H~\cite{shui2024promptnucseg}}
&\multicolumn{2}{c}{IVAAN-H\textsuperscript{*}~\cite{jeong2026ivaan}}
&\multicolumn{2}{c}{UN-SAM\textsuperscript{*}~\cite{chen2025unsam}}
&\multicolumn{2}{c}{TCINet}\\
Tissue &mPQ&bPQ&mPQ&bPQ&mPQ&bPQ&mPQ&bPQ&mPQ&bPQ&mPQ&bPQ\\
\midrule
Adrenal Gland & 0.513 & 0.709 & 0.508 & 0.715 & 0.513 & 0.723 & 0.587 & 0.736 & 0.518 & 0.726 & 0.527 & 0.720 \\
Bile Duct & 0.489 & 0.678 & 0.516 & 0.696 & 0.501 & 0.698 & 0.532 & 0.715 & 0.523 & 0.731 & 0.529 & 0.699 \\
Bladder & 0.584 & 0.707 & 0.601 & 0.714 & 0.604 & 0.721 & 0.615 & 0.735 & 0.606 & 0.723 & 0.611 & 0.731 \\
Breast & 0.518 & 0.675 & 0.514 & 0.723 & 0.532 & 0.684 & 0.550 & 0.684 & 0.535 & 0.686 & 0.537 & 0.689 \\
Cervix & 0.498 & 0.687 & 0.502 & 0.695 & 0.512 & 0.698 & 0.591 & 0.714 & 0.521 & 0.699 & 0.533 & 0.699 \\
Colon & 0.449 & 0.592 & 0.464 & 0.601 & 0.469 & 0.610 & 0.481 & 0.611 & 0.494 & 0.643 & 0.477 & 0.617 \\
Esophagus & 0.545 & 0.668 & 0.559 & 0.674 & 0.571 & 0.692 & 0.585 & 0.695 & 0.583 & 0.724 & 0.560 & 0.682 \\
Head \& Neck & 0.491 & 0.654 & 0.528 & 0.672 & 0.510 & 0.669 & 0.511 & 0.681 & 0.511 & 0.670 & 0.513 & 0.672 \\
Kidney & 0.537 & 0.709 & 0.574 & 0.728 & 0.579 & 0.712 & 0.554 & 0.680 & 0.587 & 0.738 & 0.558 & 0.721 \\
Liver & 0.522 & 0.732 & 0.535 & 0.743 & 0.533 & 0.737 & 0.566 & 0.737 & 0.534 & 0.738 & 0.547 & 0.746 \\
Lung & 0.431 & 0.643 & 0.442 & 0.646 & 0.440 & 0.658 & 0.483 & 0.647 & 0.452 & 0.673 & 0.427 & 0.655 \\
Ovarian & 0.539 & 0.672 & 0.531 & 0.668 & 0.544 & 0.686 & 0.574 & 0.693 & 0.558 & 0.706 & 0.543 & 0.671 \\
Pancreatic & 0.472 & 0.666 & 0.480 & 0.673 & 0.497 & 0.686 & 0.540 & 0.678 & 0.513 & 0.739 & 0.521 & 0.689 \\
Prostate & 0.532 & 0.682 & 0.532 & 0.678 & 0.546 & 0.698 & 0.568 & 0.707 & 0.548 & 0.721 & 0.563 & 0.695 \\
Skin & 0.434 & 0.657 & 0.422 & 0.666 & 0.411 & 0.661 & 0.450 & 0.695 & 0.434 & 0.660 & 0.436 & 0.668 \\
Stomach & 0.471 & 0.702 & 0.451 & 0.706 & 0.456 & 0.712 & 0.548 & 0.705 & 0.475 & 0.727 & 0.471 & 0.721 \\
Testis & 0.513 & 0.696 & 0.509 & 0.698 & 0.547 & 0.715 & 0.579 & 0.720 & 0.578 & 0.734 & 0.528 & 0.710 \\
Thyroid & 0.452 & 0.715 & 0.470 & 0.704 & 0.472 & 0.722 & 0.536 & 0.740 & 0.483 & 0.719 & 0.489 & 0.722 \\
Uterus & 0.474 & 0.663 & 0.517 & 0.662 & 0.495 & 0.674 & 0.526 & 0.682 & 0.506 & 0.683 & 0.474 & 0.650 \\
\midrule
Average & 0.498 & 0.679 & 0.508 & 0.685 & 0.512 & 0.692 & 0.546 & 0.698 & 0.524 & 0.707 & 0.519 & 0.692 \\
Std.~dev. & 0.041 & 0.032 & --- & --- & 0.009 & 0.015 & --- & --- & --- & --- & 0.010 & 0.006 \\
\bottomrule
\end{tabular*}
\end{table*}

\subsection{Tissue-Conditioned Cross-Branch Interaction (TC-TACBI)}
\label{sec:tacbi}

TC-TACBI routes information across three asymmetric paths using
128-channel features $f^{\rm NP}$, $f^{\rm HV}$, and $f^{\rm NT}$.
A two-layer multilayer perceptron (MLP) maps tissue probabilities to
three scaling scalars (19--64--3):
\begin{equation}\label{eq:routing}
    [\alpha_{\text{NP}}, \alpha_{\text{HV}}, \alpha_{\text{NT}}]
    = \text{MLP}(\text{softmax}(\mathbf{t})).
\end{equation}
Softmax is applied to the tissue logits before the MLP
(Eq.~\ref{eq:routing}). The three output scalars have no softmax or
sigmoid constraint, so they may be positive or negative and need not
sum to one. The final-layer weights and biases are zero-initialized,
making the module an identity at initialization.

\textbf{Paths 1 and 2 (spatial).} The HV$\to$NP path uses distance
features to guide foreground separation. In the reverse direction,
NP$\to$HV uses foreground features to guide distance regression.
Each source branch generates a one-channel mask with two depthwise
$3\times3$ convolutions, BN/GELU after the first, a $1\times1$
projection, and sigmoid. Target branch $k$ is updated as
$f^k+\alpha_k(f^k\odot M_k)$, where $\odot$ denotes element-wise
multiplication. The target retains its own representation; the
source controls where it is reweighted.

\textbf{Path 3 (channel).} For NT, global average pooling (GAP)
summarizes NP and HV into two 128-dimensional vectors. Their
concatenation passes through a 256--256--128 MLP and sigmoid to
form a channel mask, followed by the same residual update.
The NP and HV branches share pixel-level geometry and guide each
other spatially, while both supply morphological context to the NT
branch through global descriptors. All masks are computed from the
original features, allowing parallel updates.

The 1,475-parameter controller accepts the full tissue probability
vector, without hard assignments or tissue-specific parameter branches.
Image features determine the masks; tissue coefficients adjust their
strength, leaving the interaction directions unchanged. Tissue labels
supervise training, but the image alone supplies features and tissue
probabilities at inference.

\subsection{Heads and Supervision}
For NP and HV, we use an LKCell-inspired block~\cite{cui2024lkcell}
with a $13\times13$ depthwise--pointwise block followed by a
$3\times3$ depthwise--pointwise block at dilation 3. A residual
shortcut preserves the input as the convolutions add spatial context.
The NT head uses a convolution. NP supervision combines focal
Tversky~\cite{salehi2017tversky,lin2017focal}, Dice, and boundary-weighted
binary cross-entropy (BCE). HV uses mean-squared error (MSE) and
mean-squared gradient error; NT uses multiclass focal Tversky,
Dice, and cross-entropy. We also apply cross-entropy to tissue
classification.

\begin{figure*}[!t]
\centering
    \includegraphics[width=0.85\textwidth]{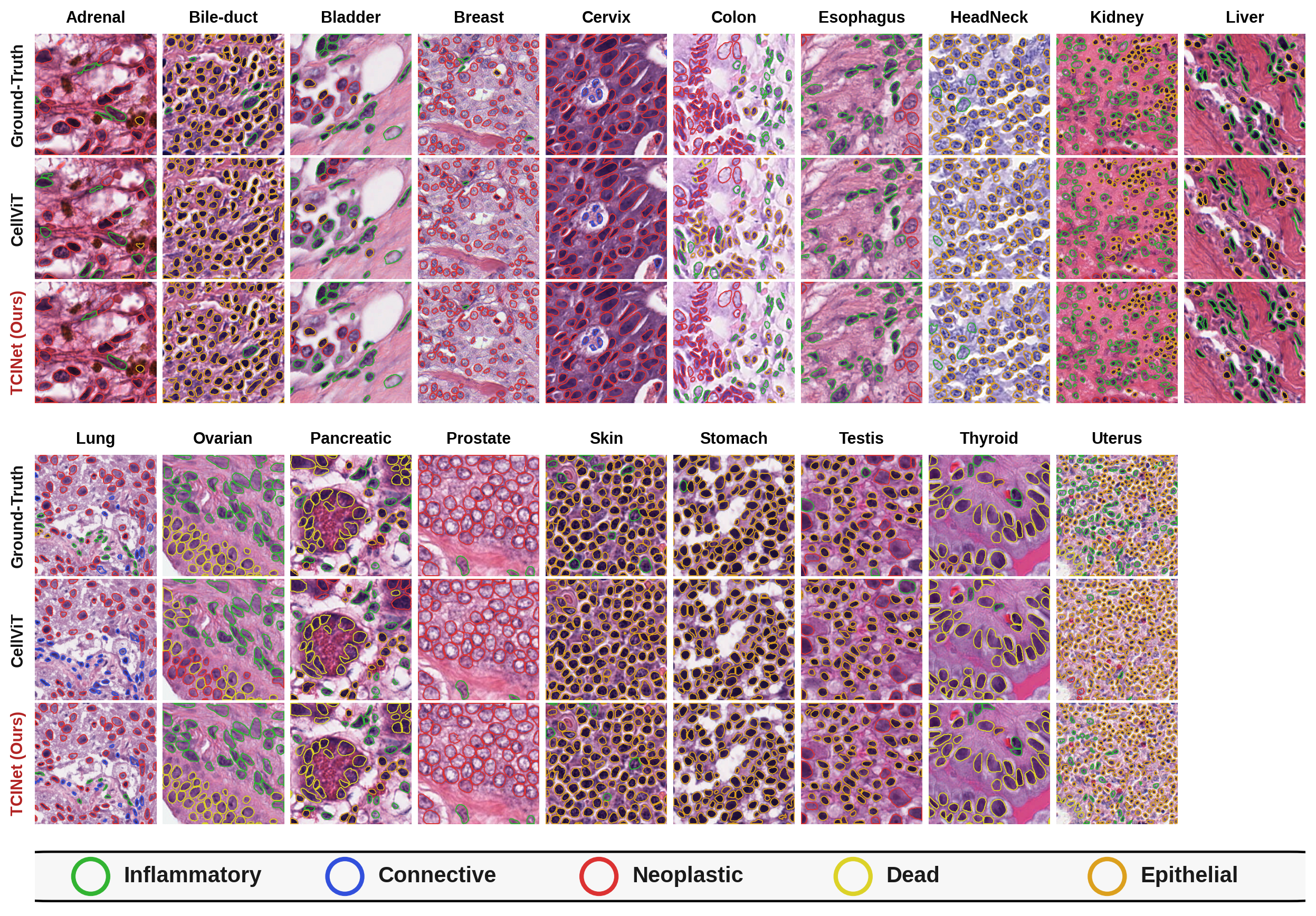}
    \caption{Qualitative comparison on PanNuke across 19 tissue
    types. For each tissue, from top to bottom: ground truth,
    CellViT-SAM-H prediction, and TCINet prediction. Nuclei contours:
    \textcolor{red}{Neoplastic}, \textcolor{orange}{Epithelial},
    \textcolor{green}{Inflammatory}, \textcolor{blue}{Connective},
    \textcolor{yellow}{Dead}.}
    \label{fig:visualization}
\end{figure*}

\begin{figure*}[!t]
\centering
    \includegraphics[width=0.85\textwidth]{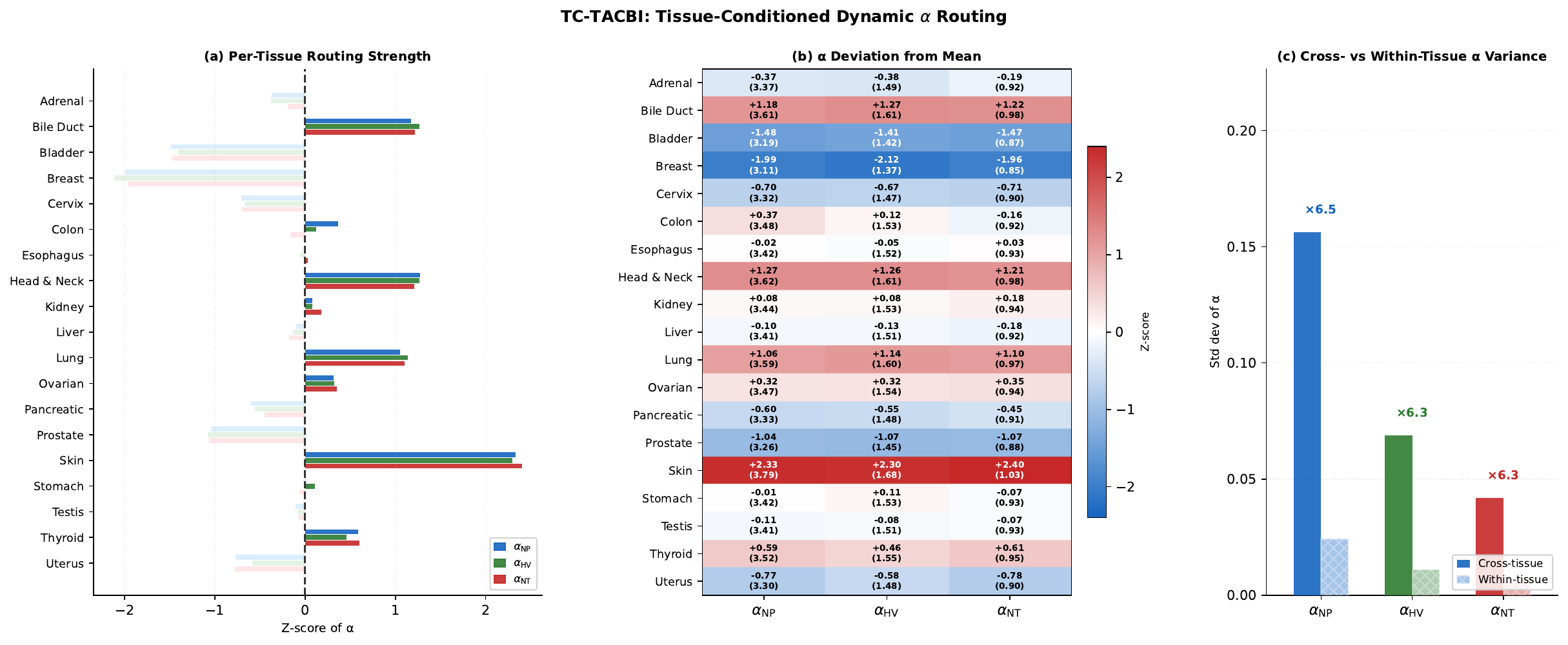}
    \caption{Tissue-conditioned attention modulation. \textbf{(a,b)}~Per-tissue
    mean $\alpha$, z-scored across tissues for each route; parentheses
    in (b) show raw means. Distinct tissue profiles indicate different
    modulation strengths for the three attention paths. \textbf{(c)}~Across-tissue
    standard deviation of tissue means versus mean within-tissue
    standard deviation, with ratios of 6.300--6.500. Bars indicate
    standard deviations, not variances.}
    \label{fig:alpha}
\end{figure*}

\section{Experiments}
\label{sec:experiments}

\subsection{Datasets, Metrics, and Implementation}
We evaluate TCINet on PanNuke~\cite{gamper2020pannuke}, containing
7,904 image patches from 19 tissue types annotated for five nucleus
classes. Evaluation uses panoptic quality~\cite{kirillov2019panoptic},
reporting mPQ and bPQ averaged across tissue types under the official
three-fold protocol, together with detection $F_{1,d}$. Zero-shot tests
on MoNuSeg~\cite{kumar2020monuseg} evaluate transfer without
target-domain fine-tuning. TCINet is trained for 130 epochs using
AdamW~\cite{loshchilov2019adamw}, a learning rate of $3\times10^{-4}$,
and weight decay of $10^{-4}$. The encoder is frozen for the first
25 epochs. Training uses color jittering and elastic deformations.
For instance extraction, we threshold NP at 0.500, remove components
below 10 pixels, and use HV gradients for seeded watershed.
A majority vote over the NT map assigns each instance its type.

\subsection{Comparison with State-of-the-Art}

Table~\ref{tab:pannuke_hierarchical} compares paper-reported PanNuke
results. TCINet achieves mPQ~=~0.519 and bPQ~=~0.692, surpassing
CellViT-SAM-H and LKCell-L on both metrics. Compared to the two-stage
PromptNucSeg-H (mPQ~0.512, bPQ~0.692), TCINet improves mPQ while
matching bPQ at the reported precision in a single-stage network.
IVAAN-H and UN-SAM report higher average scores on both metrics;
their additional text and domain conditioning are noted in the table.

\textbf{Per-category PQ.} TCINet scores higher than PromptNucSeg-H
in four of the five classes (Table~\ref{tab:nuclei_type_comparison}).
Dead PQ rises from 0.161 to 0.197; Neoplastic PQ is slightly lower,
at 0.594 versus 0.598. For Epithelial nuclei, TCINet reaches 0.600,
the highest value among the listed methods, compared with IVAAN-H's
0.591. 

\begin{table}[H]
\centering\scriptsize
\setlength{\tabcolsep}{4pt}
\caption{PanNuke PQ by nucleus type. Neo.: Neoplastic; Infl.:
Inflammatory; Conn.: Connective; Epit.: Epithelial. Baseline
values are paper-reported three-fold scores.
$^{\dagger}$TSFD-Net uses a non-official split.
$^{*}$Competitive recent efforts: IVAAN-H uses pretrained CLIP
text features and attribute-guided alignment.}
\label{tab:nuclei_type_comparison}
\begin{tabular}{lccccc}
\toprule
Method & Neo. & Infl. & Conn. & Dead & Epit. \\
\midrule
HoVer-Net~\cite{graham2019hovernet}
  & 0.551 & 0.417 & 0.388 & 0.139 & 0.491 \\
TSFD-Net\textsuperscript{\ensuremath{\dagger}}~\cite{ilyas2022tsfd}
  & 0.557 & 0.383 & 0.404 & 0.142 & 0.536 \\
CellViT-256~\cite{horst2024cellvit}
  & 0.567 & 0.405 & 0.405 & 0.144 & 0.559 \\
CellViT-SAM-H~\cite{horst2024cellvit}
  & 0.581 & 0.417 & 0.423 & 0.149 & 0.583 \\
LKCell-B~\cite{cui2024lkcell}
  & 0.585 & 0.440 & 0.414 & 0.144 & 0.579 \\
LKCell-L~\cite{cui2024lkcell}
  & 0.586 & 0.438 & 0.417 & 0.172 & 0.584 \\
PromptNucSeg-H~\cite{shui2024promptnucseg}
  & 0.598 & 0.441 & 0.433 & 0.161 & 0.582 \\
IVAAN-H\textsuperscript{*}~\cite{jeong2026ivaan}
  & 0.605 & 0.481 & 0.452 & 0.204 & 0.591 \\
\midrule
\textbf{TCINet (Ours)}
  & 0.594 & 0.454 & 0.444 & 0.197 & 0.600 \\
\bottomrule
\end{tabular}
\end{table}

\textbf{Model cost.} Our TCINet and CellViT-SAM-H implementations
contain 700.710 M and 699.740 M parameters, respectively, a 0.140\%
increase. For a $256\times256$ input at batch size 1, PyTorch
operator counting estimates 768.500 G floating-point operations
(FLOPs) for TCINet and 860.400 G for the baseline, a 10.700\% reduction.
We count two FLOPs per multiply--add and exclude preprocessing
and post-processing. Shared decoder computation allows TCINet to
add task adaptation and interaction with little net parameter growth.

\begin{samepage}
\textbf{MoNuSeg generalization.} Without fine-tuning, TCINet obtains
$F_{1,d}=0.876$ using 256-pixel tiles with 64-pixel overlap,
compared with CellViT-SAM-H's reported 0.868
(Table~\ref{tab:monuseg}). Recall increases from 0.893 to 0.908.\par
\end{samepage}

\begin{table}[H]
\centering\scriptsize
\caption{MoNuSeg zero-shot transfer using 256-pixel tiles with
64-pixel overlap. $P_d$, $R_d$: detection precision and recall.
Baseline values are from~\cite{horst2024cellvit}.}
\label{tab:monuseg}
\begin{tabular}{lcccc}\toprule
Method & bPQ & $P_d$ & $R_d$ & $F_{1,d}$\\\midrule
CellViT-256 & 0.660 & 0.841 & 0.886 & 0.863\\
CellViT-SAM-H & 0.671 & 0.846 & 0.893 & 0.868\\
TCINet & 0.672 & 0.848 & 0.908 & 0.876\\
\bottomrule\end{tabular}
\end{table}

\subsection{Ablation Study}

Table~\ref{tab:ablation} evaluates sequential component additions on
fold 0, starting from a reimplemented CellViT-SAM-H. Adding
LKCellBlock and boundary loss increases mPQ by 0.003. TC-TACBI
contributes the largest gain, raising mPQ from 0.505 to 0.525.
TSFA adds a further 0.003, showing measurable value in adapting
skip-connection features per branch when interaction is already enabled.

\begin{table}[H]
\centering
\small
\setlength{\tabcolsep}{3pt}
\caption{Sequential component ablation on PanNuke fold 0.
Baseline: reimplemented CellViT-SAM-H. LK: LKCellBlock;
BL: boundary-weighted loss. Each row adds components to the
preceding configuration.}
\label{tab:ablation}
\begin{tabular}{l cccc ccc}
\toprule
\textbf{Model} & \textbf{LK} & \textbf{BL} & \shortstack{TC-\\TACBI}
  & \textbf{TSFA}
  & \textbf{mPQ}$\uparrow$ & \textbf{bPQ}$\uparrow$
  & \textbf{$F_{1,d}$}$\uparrow$ \\
\midrule
Baseline     & & & & & 0.502 & 0.683 & 0.840 \\
+LK+BL       & \checkmark & \checkmark & & & 0.505 & 0.686 & 0.840 \\
+TC-TACBI    & \checkmark & \checkmark & \checkmark & & 0.525 & 0.693 & 0.844 \\
TCINet~(Ours) & \checkmark & \checkmark & \checkmark & \checkmark
  & \textbf{0.528} & \textbf{0.696} & \textbf{0.847} \\
\bottomrule
\end{tabular}
\end{table}

\subsection{Qualitative Results}
Fig.~\ref{fig:visualization} shows CellViT-SAM-H and TCINet
predictions alongside ground truth across 19 tissue types.
Fig.~\ref{fig:alpha} shows distinct coefficient profiles across
tissues. In the sampled patches, variation between tissue means
exceeds variation within each tissue.

\section{Discussion and Limitations}
TCINet adapts shared skip features per task and uses image-predicted
tissue probabilities to control three asymmetric interaction paths
in a single-stage NP/HV/NT network. Tissue conditioning is handled
by a 1,475-parameter controller at one decoder level.
IVAAN-H~\cite{jeong2026ivaan} augments a SAM visual encoder with
pretrained CLIP text features, attribute-guided contrastive supervision,
and query-based instance decoding. UN-SAM~\cite{chen2025unsam}
uses supplied domain information with domain-common and domain-specific
encoder mappings, self-prompt generation, and domain-query decoding.
These approaches emphasize vision--language alignment and
domain-adaptive prompting, respectively, while TCINet focuses on
feature specialization and interaction among prediction tasks.

The fold-0 ablation shows gains from TC-TACBI and TSFA in the
evaluated sequence. Further ablations are needed to assess cross-fold
stability and isolate tissue conditioning and individual paths.

\section{Conclusion}
We presented TCINet, a nuclei instance segmentation model built on
tissue-conditioned adaptive cross-branch interaction. TSFA adapts
shared skip-connection features per branch before TC-TACBI controls
cross-branch attention. On PanNuke, TCINet achieves mPQ~=~0.519 and
bPQ~=~0.692, with zero-shot $F_{1,d}=0.876$ on MoNuSeg. Sequential
ablation shows gains from both interaction and adaptation.

\FloatBarrier


\begin{thebibliography}{00}
\bibitem{graham2019hovernet}
S.~Graham \textit{et al.},
``HoVer-Net: Simultaneous segmentation and classification of nuclei
in multi-tissue histology images,''
\textit{Medical Image Analysis}, vol.~58, p.~101563, 2019.
\bibitem{horst2024cellvit}
F.~H\"{o}rst \textit{et al.},
``CellViT: Vision transformers for precise cell segmentation and
classification,''
\textit{Medical Image Analysis}, vol.~94, p.~103143, 2024.
\bibitem{kirillov2023sam}
A.~Kirillov \textit{et al.},
``Segment anything,''
in \textit{Proc. ICCV}, 2023, pp.~4015--4026.
\bibitem{cui2024lkcell}
Z.~Cui \textit{et al.},
``LKCell: Efficient cell nuclei instance segmentation with large
convolution kernels,''
\textit{arXiv preprint arXiv:2407.18054}, 2024.
\bibitem{shui2024promptnucseg}
Z.~Shui \textit{et al.},
``Unleashing the power of prompt-driven nucleus instance
segmentation,''
in \textit{Proc. ECCV}, 2024, pp.~288--304.
\bibitem{misra2016cross}
I.~Misra, A.~Shrivastava, A.~Gupta, and M.~Hebert,
``Cross-stitch networks for multi-task learning,''
in \textit{Proc. IEEE/CVF CVPR}, 2016, pp.~3994--4003.
\bibitem{gao2019nddr}
Y.~Gao, Q.~She, J.~Ma, M.~Zhao, W.~Liu, and A.~L.~Yuille,
``NDDR-CNN: Layerwise feature fusing in multi-task CNNs by neural
discriminative dimensionality reduction,''
in \textit{Proc. IEEE/CVF CVPR}, 2019, pp.~3205--3214.
\bibitem{hu2018squeeze}
J.~Hu, L.~Shen, and G.~Sun,
``Squeeze-and-excitation networks,''
in \textit{Proc. IEEE/CVF CVPR}, 2018, pp.~7132--7141.
\bibitem{woo2018cbam}
S.~Woo, J.~Park, J.-Y.~Lee, and I.~S.~Kweon,
``CBAM: Convolutional block attention module,''
in \textit{Proc. ECCV}, 2018, pp.~3--19.
\bibitem{gamper2020pannuke}
J.~Gamper \textit{et al.},
``PanNuke dataset extension, insights and baselines,''
\textit{arXiv preprint arXiv:2003.10778}, 2020.
\bibitem{kumar2020monuseg}
N.~Kumar \textit{et al.},
``A multi-organ nucleus segmentation challenge,''
\textit{IEEE Trans. Med. Imaging}, vol.~39, no.~5,
pp.~1380--1391, 2020.
\bibitem{jeong2026ivaan}
J.~Jeong \textit{et al.}, \emph{IVAAN: Instance-level vision-language
alignment via attribute-guided text prompts generation for nuclei
analysis}, in \textit{Proc. IEEE/CVF CVPR}, 2026, pp.~29305--29314.
\bibitem{chen2025unsam}
Z.~Chen, Q.~Xu, X.~Liu, and Y.~Yuan, \emph{UN-SAM:
Domain-adaptive self-prompt segmentation for universal nuclei
images}, \textit{Medical Image Analysis}, vol.~103,
p.~103607, 2025.
\bibitem{ilyas2022tsfd}
T.~Ilyas \textit{et al.},
``TSFD-Net: Tissue specific feature distillation network for nuclei
segmentation and classification,''
\textit{Neural Networks}, vol.~151, pp.~1--15, 2022.
\bibitem{chen2024sfe}
J.~Chen, G.~Yang, A.~Liu, X.~Chen, and J.~Liu,
``SFE-Net: Spatial-frequency enhancement network for robust nuclei
segmentation in histopathology images,''
\textit{Computers in Biology and Medicine}, vol.~171,
p.~108131, 2024.
\bibitem{perez2018film}
E.~Perez, F.~Strub, H.~de~Vries, V.~Dumoulin, and A.~Courville,
``FiLM: Visual reasoning with a general conditioning layer,''
in \textit{Proc. AAAI}, 2018, pp.~3942--3951.
\bibitem{houlsby2019adapter}
N.~Houlsby \textit{et al.},
``Parameter-efficient transfer learning for NLP,''
in \textit{Proc. ICML}, 2019, pp.~2790--2799.
\bibitem{hu2022lora}
E.~J.~Hu \textit{et al.},
``LoRA: Low-rank adaptation of large language models,''
in \textit{Proc. ICLR}, 2022.
\bibitem{chen2022vitadapter}
Z.~Chen \textit{et al.},
``Vision transformer adapter for dense predictions,''
in \textit{Proc. ICLR}, 2023.
\bibitem{wu2023medical}
J.~Wu \textit{et al.},
``Medical SAM adapter: Adapting segment anything model for medical
image segmentation,''
\textit{arXiv preprint arXiv:2304.12620}, 2023.
\bibitem{salehi2017tversky}
S.~S.~M.~Salehi, D.~Erdogmus, and A.~Gholipour,
``Tversky loss function for image segmentation using 3D fully
convolutional deep networks,''
in \textit{Proc. MICCAI Workshop on Machine Learning in Medical
Imaging}, 2017, pp.~379--387.
\bibitem{lin2017focal}
T.-Y.~Lin, P.~Goyal, R.~Girshick, K.~He, and P.~Doll\'{a}r,
``Focal loss for dense object detection,''
in \textit{Proc. IEEE/CVF ICCV}, 2017, pp.~2980--2988.
\bibitem{kirillov2019panoptic}
A.~Kirillov, K.~He, R.~Girshick, C.~Rother, and P.~Doll\'{a}r,
``Panoptic segmentation,''
in \textit{Proc. IEEE/CVF CVPR}, 2019, pp.~9404--9413.
\bibitem{loshchilov2019adamw}
I.~Loshchilov and F.~Hutter,
``Decoupled weight decay regularization,''
in \textit{Proc. ICLR}, 2019.
\end{thebibliography}
\end{document}